%% file: main.tex
\documentclass[letterpaper, 10 pt, conference]{ieeeconf}

\IEEEoverridecommandlockouts

\usepackage{etextools}
\usepackage{amssymb}
\usepackage{float}
\usepackage{makeidx}
\usepackage{amsmath}
\usepackage{bbm}
\usepackage{epsfig}
\usepackage{epsf}
\usepackage{psfrag}
\usepackage{verbatim}
\usepackage{color}
\usepackage{multirow}
\usepackage{tabularx}
\usepackage{booktabs}
\usepackage[tight,footnotesize]{subfigure}
\usepackage{array}
\usepackage{soul}
\usepackage{footnote}
\usepackage{cite}
\usepackage{dblfloatfix}

\usepackage{xcolor}
\usepackage{color, colortbl}
\usepackage[colorlinks,bookmarksnumbered,citecolor=orange,urlcolor=orange]{hyperref}
\usepackage{graphicx}
\graphicspath{{Figures}}
\DeclareGraphicsExtensions{.pdf,.png}
\usepackage{bigstrut}
\usepackage[english]{babel}

\usepackage{csquotes}

\usepackage[T1]{fontenc}
\usepackage{algorithm, setspace}
\usepackage{algpseudocode}
\usepackage{url}
\usepackage{multirow}
\usepackage{float}
\usepackage{xcolor}
\usepackage{hyperref}
 \hypersetup{
     colorlinks=true,
     linkcolor=orange,
     filecolor=orange,
     citecolor=orange,      
     urlcolor=orange,
     }

\DeclareMathOperator{\choiceh}{\Xi_{\mathcal{H}}}
\DeclareMathOperator{\choicer}{\Xi_{\mathcal{R}}}

\usepackage{balance}

\title{\LARGE

\textit{I Know What You Meant:} \\
Learning Human Objectives by (Under)estimating Their Choice Set

}

\author{Ananth Jonnavittula and Dylan P. Losey
\thanks{The authors are members of the Collaborative Robotics Lab (\href{https://collab.me.vt.edu/}{Collab}), Dept. of Mechanical Engineering, Virginia Tech, Blacksburg, VA 24061.
\newline
{e-mail: \texttt{\{ananth, losey\}@vt.edu}}}
}

\begin{document}
\maketitle

\begin{abstract}


Assistive robots have the potential to help people perform everyday tasks. 
However, these robots first need to learn what it is their user wants them to do. 
Teaching assistive robots is hard for inexperienced users, elderly users, and users living with physical disabilities, since often these individuals are unable to show the robot their desired behavior.
We know that \textit{inclusive} learners should give human teachers credit for what they cannot demonstrate.
But today's robots do the opposite: they assume \textit{every} user is capable of providing \textit{any} demonstration.
As a result, these robots learn to mimic the demonstrated behavior, even when that behavior is not what the human really meant!
Here we propose a different approach to reward learning: robots that reason about the user's demonstrations in the context of similar or simpler alternatives.
Unlike prior works --- which err towards overestimating the human's capabilities --- here we err towards \textit{underestimating} what the human can input (i.e., their choice set).
Our theoretical analysis proves that underestimating the human's choice set is risk-averse, with better worst-case performance than overestimating.
We formalize three properties to generate similar and simpler alternatives. Across simulations and a user study, our resulting algorithm better extrapolates the human's objective.
See the user study here: \url{https://youtu.be/RgbH2YULVRo}.

\end{abstract}

\smallskip





\input{intro}
\input{related}
\input{problem}
\input{theory}
\input{method}
\input{simulations}
\input{user-study}

\input{conclusions}


\newpage
\balance
\bibliographystyle{IEEEtran}
\bibliography{IEEEabrv,bibtex}

\end{document}

%% file: intro.tex
\section{Introduction}

\begin{figure*}[t]
	\begin{center}
		\includegraphics[width=1.6\columnwidth]{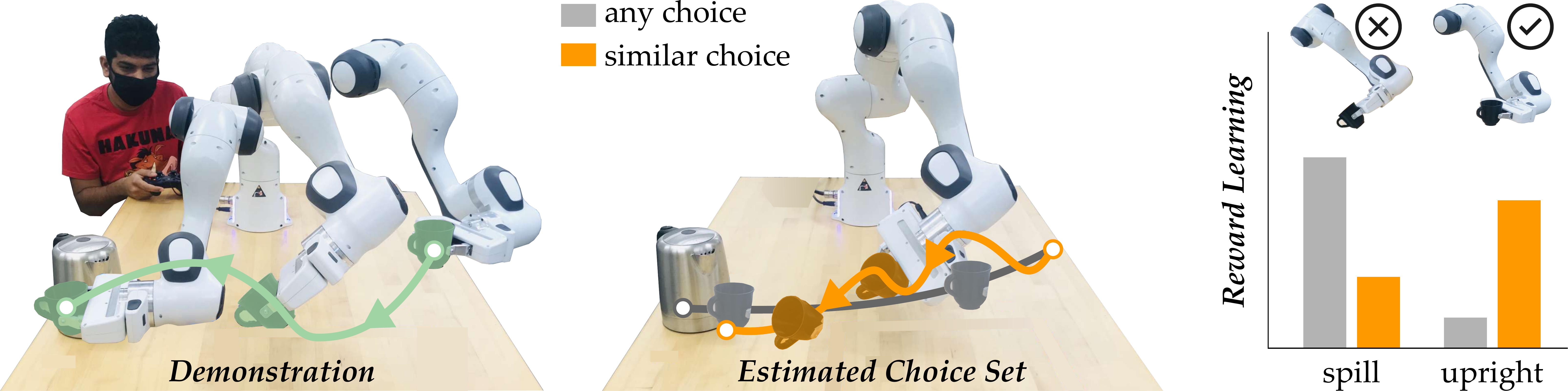}
		\vspace{-0.5em}
		\caption{(Left) User teaches an assistive robot arm to carry coffee upright. Because of their limitations, this user's best demonstration unintentionally tilts the cup. (Middle) When the robot ignores these limitations --- and compares the human's input to \textcolor{gray}{\textbf{any trajectory}} --- it learns to match the human's demonstration and also spill coffee. To make learning more inclusive, we instead compare the human's demonstration to \textcolor{orange}{\textbf{simpler and similar trajectories}}. (Right) Because these alternatives spill more coffee than the human, the robot now extrapolates what the user really wanted: keeping the coffee upright.}
		\label{fig:front}
	\end{center}
 	\vspace{-2.0em}
\end{figure*}

Imagine that you're using an assistive robot arm, and you want to teach this arm to carry a cup of coffee upright. You control the robot's motion using a joystick \cite{losey2020controlling, herlant2016assistive, argall2018autonomy}. The robot has many degrees-of-freedom, and --- as you teleoperate the arm to provide your demonstration --- you must carefully orchestrate each joint to keep the coffee upright. It's possible for a dexterous, experienced human to teleoperate this task correctly. But you're not as skilled with the control interface: \textit{because of your limitations} the best you can do is to avoid turning over the cup and spilling all the coffee (see Fig.~\ref{fig:front}).

We refer to the set of possible demonstrations that a given human can show the robot as the human's \textit{choice set} \cite{jeon2020reward, freedmanchoice}. Robots learn what the human wants by reasoning about the human's actual demonstrations in the context of this choice set. For instance, if the human's demonstrations keep the coffee more vertical than the alternatives within their choice set, it's likely that the human values keeping the coffee upright. Importantly, when we interact with robots, our choice sets are \textit{inherently constrained} by our own physical and cognitive capabilities --- and these limitations vary from one person to another \cite{evans2015learning, kahneman2013prospect, choi2014more}. Take our example: although an expert is able to show a trajectory with the coffee roughly upright, the best we can do is to keep the cup from flipping over. Uncertainty over what this choice set is makes it hard for the robot to learn: did we tilt the coffee because we meant to, or because we were unable to keep it upright?

Existing work on learning human objectives from demonstrations often ignores user limitations, and assumes that the human's choice set includes \textit{every} demonstration that is consistent with the robot's dynamics \cite{osa2018algorithmic,abbeel2004apprenticeship, ziebart2008maximum, ramachandran2007bayesian, finn2016guided}. This results in robots that learn to \textit{match} your demonstrated behavior, so that if all your demonstrations spill some coffee, the robot also learns to spill. We explore the opposite perspective:
\begin{center}\vspace{-0.4em}
\textit{Inclusive robots should assume that the human may have limitations, and can only show alternative behaviors that are} similar to \textit{or} simpler than \textit{their actual demonstrations.}\vspace{-0.4em}
\end{center}
Returning to our coffee example, now the robot narrows down its estimate of the human's choice set to only include the observed human trajectories and sparse and noisy alternatives. Compared to these alternatives, our actual demonstrations better keep the cup closer to vertical. Here the robot \textit{extrapolates} what we really meant, despite the fact that we spilled coffee in all our demonstrations (see Fig.~\ref{fig:front}).

Overall, we make the following contributions:

\noindent \textbf{Underestimating vs. Overestimating Choice Sets.} Robots will inevitably get the human's choice set wrong. We theoretically compare overestimating the human's choice set (i.e., assuming users can provide any demonstration) to underestimating the human's choice set (i.e., users can only provide a few demonstrations). We prove that underestimating is a risk-averse approach with better worst-case learning.

\noindent \textbf{Generating Choice Sets from Demonstrations.} How does the robot get its estimate of a human's choice set? Our analysis finds that waiting for human teachers to demonstrate their choice set is intractable. To address this, we formalize three properties that robots can leverage to generate trajectories that are similar to or simpler than the human's actual choices.

\noindent \textbf{Conducting a User Study.} We compare our inclusive approach to state-of-the-art baselines. Our results suggest that robots which err towards underestimating the human's choice set better extrapolate the human's underlying objective.

%% file: related.tex
\section{Related Work}

\noindent \textbf{Application -- Assistive Robots.} Assistive robots, such as wheelchair-mounted robot arms, promise to improve the autonomy and independence of users living with physical disabilities \cite{taylor2018americans}. However, people's ability to control assistive robots is often restricted, both by their own physical impairments and the teleoperation interfaces they leverage \cite{argall2018autonomy, losey2020controlling, gopinath2020customized, herlant2016assistive, aronson2018eye, muelling2017autonomy}. Expecting all users to produce the same demonstrations is unfair \cite{losey2019enabling}. We therefore develop a \textit{personalized} approach, where the robot learns from each user by placing their demonstrations in the context of similar behaviors.

\smallskip

\noindent \textbf{Learning Rewards from Demonstrations.} Prior work studies how assistive arms and other robots can learn from human demonstrations. Inverse reinforcement learning (IRL) infers what the user wants --- i.e., their underlying reward function --- from their demonstrated trajectories. To do this, the robot searches for the reward function that makes the human's trajectories appear roughly optimal \cite{osa2018algorithmic}. Crucially, today's methods assume that the human is optimizing over all possible trajectories; accordingly, when human demonstrates suboptimal behavior, the robot thinks this behavior is better than every other option, and learns the reward function that best reproduces what the human demonstrated \cite{abbeel2004apprenticeship, ziebart2008maximum, ramachandran2007bayesian, finn2016guided}.

Recent research explores how this learning model \textit{misses out} on the way humans actually behave \cite{bajcsy2018learning, reddy2018you, bobu2020quantifying}. Most relevant are \cite{jeon2020reward} --- where the authors formalize choice sets in reward learning --- and \cite{freedmanchoice} --- where the authors experimentally compare different classes of choice set misspecification. We also build upon inverse reinforcement learning, but develop a formalism for generating smaller, constrained choice sets from the human's demonstrations.

\smallskip

\noindent \textbf{Outperforming Imperfect Demonstrations.} Our goal in generating these choice sets is for robots to learn rewards which better align with what the human wants than the human's demonstrations. Other works outperform imperfect and suboptimal demonstrations by eliciting additional types of human feedback \cite{grollman2011donut, shiarlis2016inverse,biyik2020learning, sadigh2017active, choi2019robust}. We do not gather additional feedback; instead, our approach is most similar to \cite{brown2020better, kalakrishnan2013learning, boularias2011relative}. Here the robot considers noisy perturbations of the human's original demonstrations, and extrapolates a reward function which maximizes the difference between the human's demonstrations and these noisy alternatives. Viewed within our formalism, the alternative demonstrations form the human's choice set, and leveraging noisy perturbations becomes \textit{one instance} of generating this set.

%% file: problem.tex
\section{Problem Statement}

Let's return to our motivating example, where we are teleoperating an assistive robot arm, and want to teach this arm to hold coffee upright. Each time we guide the robot through the process of picking up, carrying, and putting down our coffee cup, we show the robot a trajectory $\xi \in \choiceh$. Here $\xi$ is a sequence of robot state-action pairs, and $\choiceh$ is our choice set: i.e., the set of all trajectories \textit{we are capable} of showing to the robot. Of course, the robot does not know what our choice set is --- we could be an expert user, capable of keeping the cup perfectly vertical, or an inexperienced user, who struggles to keep the cup from flipping over. The robot therefore works with $\choicer$, an \textit{estimate} of $\choiceh$.

Over repeated interactions we show the robot $N$ trajectories from our choice set. Given all of these user demonstrations $\mathcal{D} = \{\xi_1, \xi_2, \ldots, \xi_N\}$, the robot recovers our reward function $r_{\theta}(\xi) \rightarrow \mathbb{R}$, i.e., what we want the robot to optimize for across a trajectory\footnote{In our experiments we are consistent with previous work on inverse reinforcement learning and use $r_{\theta}(\xi) = \theta \cdot \Phi(\xi)$, where $\Phi(\xi) \in \mathbb{R}^k$ are known features and $\theta \in \mathbb{R}^k$ are unknown reward weights \cite{osa2018algorithmic, abbeel2004apprenticeship, ziebart2008maximum, jeon2020reward}.}. More formally, the robot tries to infer $r_{\theta}$ (what we value) given $\mathcal{D}$ (the trajectories we have shown) and $\choicer$ (the trajectories the robot thinks we can show).

\smallskip

\noindent \textbf{Bayesian Inference.} Let belief $b$ denote the robot's learned probability distribution over $r_{\theta}$. Applying Bayes' Theorem:
\begin{equation} \label{eq:P1}
    b(r_{\theta}) = P(r_{\theta} \mid \mathcal{D}, \choicer) \propto P(\mathcal{D} \mid r_{\theta}, \choicer) \cdot P(r_{\theta})
\end{equation}
where $P(r_{\theta})$ is the robot's prior over the human's reward. Since we assume the human's demonstrations are conditionally independent \cite{osa2018algorithmic, biyik2020learning}, Equation (\ref{eq:P1}) simplifies to:
\begin{equation} \label{eq:P2}
    b(r_{\theta}) \propto P(r_{\theta}) \prod_{\xi \in \mathcal{D}}  P(\xi \mid r_{\theta}, \choicer)
\end{equation}
The crucial term is $P(\xi \mid r_{\theta}, \choicer)$, the likelihood of observing $\xi$ given that the human has reward $r_{\theta}$ and choice set $\choicer$. Within robot learning \cite{ziebart2008maximum} and cognitive science \cite{baker2009action}, this is commonly written using a Boltzmann-rational model:
\begin{equation} \label{eq:P3}
    P(\xi \mid r_{\theta}, \choicer) = \frac{\exp{(\beta \cdot r_{\theta}(\xi))}}{\sum_{\xi' \in \choicer}\exp{(\beta \cdot r_{\theta}(\xi'))}}
\end{equation}
Intuitively, this model asserts that a rational human picks the trajectory from their choice set that noisily maximizes their reward, and the hyperparameter $\beta \geq 0$ captures how close-to-rational the human is. Substituting this likelihood function into Equation (\ref{eq:P2}), we arrive at the \textit{robot's learning rule}:
\begin{equation} \label{eq:P4}
    b(r_{\theta}) \propto \frac{\exp{(\beta \cdot \sum_{\xi \in \mathcal{D}}r_{\theta}(\xi))}}{\big(\sum_{\xi' \in \choicer}\exp{(\beta \cdot r_{\theta}(\xi'))}\big)^N} \cdot P(r_{\theta})
\end{equation}

\smallskip

\noindent \textbf{Importance of Choice Sets.} When the robot misunderstands the human's choice set (i.e, $\choicer \neq \choiceh$) this learning rule may spectacularly fail. Imagine that the robot from Fig.~\ref{fig:front} has a uniform prior over two opposite rewards that optimize for staying upright ($r_{upright}$) or spilling coffee ($r_{spill}$). As an inexperienced user the best we can do is to spill at least $75\%$ of the coffee, and we demonstrate this $\xi$ to the robot --- hoping the robot will recognize our limitations and learn $r_{upright}$. However, the robot overestimates our capabilities. Let $\choicer$ include a full spectrum of trajectories, from spilling the entire cup to spilling none at all: importantly, our $\xi$ pours more coffee than most of these alternatives in $\choicer$. Hence, if we apply Equation (\ref{eq:P4}) with this choice set, for any $\beta > 0$ the robot believes $r_{spill}$ is actually what we wanted to optimize. The robot gets it wrong here because it \textit{incorrectly estimates} what trajectories the human can input.

%% file: theory.tex
\section{Theoretical Analysis}

The problem is that robots will \textit{inevitably} get the human's choice set wrong. This is especially true in assistive robotics, where physical and control limitations vary greatly from one user to another \cite{argall2018autonomy}. If we are doomed to mistake $\choiceh$, on which side should we err? One option is to \textit{overestimate} the choices that the human can make, such that $\choicer \supseteq \choiceh$. This is traditionally done in inverse reinforcement learning, where $\choicer$ includes all feasible trajectories \cite{abbeel2004apprenticeship, ziebart2008maximum, ramachandran2007bayesian, finn2016guided}. By contrast, our insight is to err towards \textit{underestimating} the human's capabilities, such that $\choicer \subseteq \choiceh$. In this section we analyze the pros and cons of each approach.

\smallskip

\noindent \textbf{Assumptions.} To better understand how the system behaves at the limit, we assume that the human is perfectly rational, and always inputs the $\xi \in \choiceh$ that maximizes their reward (i.e., $\beta \rightarrow \infty$). We further assume that the human's choice $\xi$ is an element of $\choicer$: because the robot observes $\xi$ prior to learning, it is trivial to add $\xi$ to $\choicer$ if it is not already included. Finally, for the sake of clarity, we assume that the mapping from trajectories $\xi$ to rewards $r_{\theta}(\xi)$ is injective.

In practice, putting these assumptions together with Equation (\ref{eq:P3}) implies that either $P(\xi \mid r_{\theta}, \choicer) = 1$ (if $\xi \in \choicer$ maximizes $r_{\theta}$) or $P(\xi \mid r_{\theta}, \choicer) = 0$ (if $\xi \in \choicer$ does not maximize $r_{\theta}$). These assumptions are purely for analysis, and are removed in our simulations and user study.

\subsection{Risk-Sensitivity and Learning}

In the context of learning with Equation (\ref{eq:P4}), one definition of risk relates to \textit{confidence}. A risk-seeking robot becomes confident in its estimate of $r_{\theta}$ after just a few demonstrations, while a risk-averse robot remains uncertain even when many demonstrations are available. We quantify this confidence as the Shannon entropy over the robot's belief $b$ \cite{cover2012elements}.

But how confident should the robot actually be? Our point of reference is an ideal leaner that knows $\choiceh$. This is the gold standard, because if we knew $\choiceh$, we'd always have the right context for learning from the human's demonstrations. We therefore define risk as the difference between the robot's \textit{actual} entropy and the \textit{gold standard} entropy. A risk-seeking robot has less entropy than the gold standard (indicating it is overly confident in its estimate) while a risk-averse robot has more entropy than the gold standard (indicating it is being unnecessarily conservative).

We find that how we estimate the choice set affects risk:

\smallskip

\noindent \textbf{Proposition 1.} \textit{Robots which} overestimate \textit{the human's capabilities are risk-seeking, while robots which} underestimate \textit{the human's choice set are risk-averse}.

\smallskip

\noindent \textit{Proof.} Let $\xi \in \mathcal{D}$ be an observed human demonstration. When the robot overestimates the human's capabilities, $\choicer \supseteq \choiceh$. This increases the denominator of Equation (\ref{eq:P3}), so that $P(\xi \mid r_{\theta}, \choicer) \leq P(\xi \mid r_{\theta}, \choiceh)$ for all $r$. But from our assumptions we know that $P(\xi \mid r_{\theta}, \choicer)$ is either $0$ or $1$. Thus, $\choicer \supseteq \choiceh$ concentrates the robot's belief $b$ around the same or fewer rewards than if $\choicer = \choiceh$.

Conversely, if the robot underestimates the human's capabilities, $\choicer \subseteq \choiceh$. Removing choices decreases the denominator of Equation (\ref{eq:P3}), so that $P(\xi \mid r_{\theta}, \choicer) \geq P(\xi \mid r_{\theta}, \choiceh)$ for all $r_{\theta}$. Hence, $\choicer \subseteq \choiceh$ distributes the robot's belief $b$ around the same or more rewards than if $\choicer = \choiceh$. \hfill $\square$

\smallskip

To see this in action let's return to our motivating example, where we input a trajectory which spills some coffee. If the robot overestimates our capabilities, the robot becomes confident that we are optimizing for spilling, since there are better trajectories to choose if we preferred to keep the cup upright. But a robot that underestimates our capabilities is not as quick to eliminate other explanations. This robot realizes that the trajectory we have shown is our best choice if we intend to spill some coffee, as well as our best choice for spilling no coffee at all. Hence, the risk-averse robot thinks both of these rewards are still likely.

\subsection{Worst-Case Learning}

Next let's look at the worst that could happen when we get the human's choice set wrong. We'll think about this error in terms of Equation~(\ref{eq:P4}), where the robot learns the likelihood of each reward. In the best case the robot learns the human's true reward $r_{\theta}$, so that $b(r_{\theta}) = 1$.

The worst-case depends on how we estimate $\choiceh$:

\smallskip

\noindent \textbf{Proposition 2.} \textit{In the worst case, robots that} overestimate \textit{the human's capabilities learn the wrong reward}.

\smallskip

\noindent \textit{Proof.} The rational human chooses $\xi \in \choiceh$ to maximize their reward $r_{\theta}$. But $\choicer \supset \choiceh$, and in $\choicer$ there might be another trajectory $\xi'$ where $r_{\theta}(\xi') > r_{\theta}(\xi)$. So the human's choice $\xi$ does not maximize reward for $r_{\theta}$; instead, $\xi \in \choicer$ is the best choice for reward $r_{\theta}'$. Here $b(r_{\theta}') = 1$ and $b(r_{\theta}) = 0$. \hfill $\square$

\smallskip

\noindent \textbf{Proposition 3.} \textit{In the worst case, robots that} underestimate \textit{the human's capabilities learn nothing from demonstrations}.

\smallskip

\noindent \textit{Proof.} The rational human chooses $\xi \in \choiceh$ to maximize $r_{\theta}$. But $\choicer \subset \choiceh$ only contains this single choice, $\choicer = \{\xi\}$. So while $\xi \in \choicer$ is the best choice for reward $r_{\theta}$, it is also the best choice for all other rewards. Now ${P(\xi \mid r_{\theta}, \choicer) = 1}$ for every reward, and $b(r_{\theta}) = P(r_{\theta})$. \hfill $\square$

\smallskip

We can intuitively connect this worst-case performance to our earlier analysis of risk-sensitivity. Robots that overestimate the human's capabilities are risk-seeking, and quickly become overconfident in the reward they have learned. When these risky robots get it wrong --- for instance, thinking we want to spill the coffee --- they \textit{commit} to their mistakes, resulting in complete confidence in the wrong reward. Robots that underestimate the human's choice set err in the opposite direction. These risk-averse robots play it safe, and maintain several possible explanations for the human's behavior. When these conservative robots are overly \textit{cautious} --- e.g., severely underestimating our capabilities --- they ignore the information our demonstrations actually contain.

\subsection{Can We Rely on Human Teachers?}

To avoid this worst-case performance, one option is to rely on the human \textit{to show the robot} their choice set. Every time the human teleoperates the robot along a trajectory $\xi$, they are showing the robot another element of $\choiceh$. A na\"ive robot may assume that --- given enough time --- the human will demonstrate \textit{all} the trajectories they are capable of inputting. Using this wait-and-see approach, the robot sets $\choicer = \mathcal{D}$.

The problem here is that the user isn't picking $\xi$ to convey their choice set; instead, the human is inputting trajectories to teach the robot. Referring back to Equation~(\ref{eq:P3}), the human chooses $\xi \in \choiceh$ to noisily maximize their reward $r$. When users follow this Boltzmann-rational model, passively waiting for the user to show a diverse set of choices from $\choiceh$ becomes prohibitively time consuming:

\smallskip

\noindent \textbf{Proposition 4.} \textit{If we normalize the rewards over $\choiceh$ between $0$ and $1$, the probability that the human will show a minimal reward trajectory $\xi$ in $N$ demonstrations is bounded by:}
\begin{equation} \label{eq:T1}
    P(\xi \in \mathcal{D}) \leq 1- \Bigg[\frac{|\choiceh| - 2 + \exp{\beta}}{|\choiceh| - 1 + \exp{\beta}}\Bigg]^N
\end{equation}

\noindent \textit{Proof.} 
We seek to maximize Equation~(\ref{eq:P3}) when $r_{\theta}(\xi) = 0$ and there exists at least one $\xi'$ where $r_{\theta}(\xi') = 1$. To do this we minimize the denominator, which occurs when $r_{\theta}(\xi') = 1$ only once and $0$ otherwise. Now $P(\xi \mid r_{\theta}, \choiceh) = 1/(\exp{\beta} + |\choiceh| - 1)$, and the probability of not picking $\xi$ during the current interaction is $1 - P(\xi \mid r_{\theta}, \choiceh)$. \hfill $\square$

\smallskip

Jumping back to the motivating example, let's say there are $|\choiceh| = 2$ different types of trajectories we can demonstrate: either tilting the mug or completely flipping it over. For a Boltzmann-rational human teacher with $\beta = 5$, \textit{even if the robot waits for $50$ demonstrations}, there is at most a $29\%$ likelihood that the human demonstrates flipping the cup over. Hence, when humans are optimizing for teaching, we \textit{cannot rely} on them to convey their choice set to the robot.

%% file: method.tex
\section{Generating the Choice Set}

The last section explored how underestimating or overestimating the human's choice set affects reward learning --- but how does the robot estimate this choice set in the first place? Although we can't rely on the human to show us the entire $\choiceh$ (see Proposition 4), we can still leverage the choices the human makes. To estimate $\choiceh$ from an individual user's demonstrations, we apply our insight: $\choiceh$ only includes trajectories that are \textit{similar to} or \textit{simpler than} $\mathcal{D}$. Unlike prior work, this approach errs towards \textit{underestimating} (i.e., $\choicer \subseteq \choiceh$), since we are implicitly assuming that the user's demonstrations are the limits of their capabilities.

In what follows we formalize three properties that result in similar and simpler trajectories. These properties are not meant to be an exhaustive list, but rather a starting point for generating intuitive and inclusive choice sets.

\smallskip

\noindent \textbf{Noisy Deformations.} When humans teleoperate robots they make minor mistakes: e.g., unintentionally pressing the joystick up instead of right. We capture these \textit{local perturbations} as noisy trajectory deformations. Let $\sigma \sim p(\cdot)$ be a noise parameter. After sampling $\sigma$, we deform the human trajectory $\xi \in \mathcal{D}$ to get a similar alternative: $\xi' = f(\xi, \sigma)$. One concrete way to deform the human's trajectory is $\xi' = \xi + A \sigma$, where $A$ defines the deformation shape \cite{losey2017trajectory}. In practice, this produces alternate trajectories the user \textit{could have input} if they made small, stochastic changes to their inputs.

\begin{algorithm}[t]
  \caption{Inclusive reward learning from demonstrations}
  \label{alg}
  \begin{algorithmic}[1]
    \State Collect human demonstrations: $\mathcal{D} = \{\xi_1, \xi_2, \ldots, \xi_N \}$
    \State Initialize estimate of human's choice set: $\choicer = \mathcal{D}$
    \For{$\xi \in \mathcal{D}$}
        \State Generate counterfactual $\xi'$ from $\xi$
        \State Add the resulting trajectory $\xi'$ to $\choicer$
    \EndFor
    \State Update robot's belief given $\mathcal{D}$ and $\choicer$:
    \begin{equation*}
        P(r_{\theta} \mid \mathcal{D}, \choicer) \propto \frac{\exp{(\beta \cdot \sum_{\xi \in \mathcal{D}}r_{\theta}(\xi))}}{\big(\sum_{\xi' \in \choicer}\exp{(\beta \cdot r_{\theta}(\xi'))}\big)^N} \cdot P(r_{\theta})
    \end{equation*}
  \end{algorithmic}
\end{algorithm}

\noindent \textbf{Sparse Inputs.} To guide the robot throughout the process of picking up, carrying, and placing a coffee cup, the user must input a complex sequence of teleoperation commands. Each of these individual inputs requires user \textit{effort} and \textit{intention}. Hence, we hypothesize that some simpler trajectories are the result of \textit{less user oversight}. Let $u$ be a human teleoperation input, let $U = [u^1, u^2, \ldots]$ be the sequence of human inputs, and let $g(U) \rightarrow \xi'$ be the trajectory that results from input sequence $U$. Formally, this property searches for trajectories that are similar to $\xi \in \mathcal{D}$ but produced by sparser inputs:
\begin{equation} \label{eq:A1}
    \xi' = g(U^*), \quad U^* = \text{arg}\min ~ \| \xi - g(U)\|^2 + \lambda \cdot \|U\|_1^2
\end{equation}
where scalar $\lambda > 0$ determines the relative trade-off between trajectory similarity and input sparsity.

\noindent \textbf{Consistent Inputs.} Precisely controlling the robot arm often requires rapidly \textit{changing} inputs: e.g., adjusting the position of the cup, then rotating the cup upright, and then fixing its position again \cite{bajcsy2018learning}. Instead of expecting the user to make all of these different commands, we recognize that it is easier to \textit{maintain consistent inputs}. Using the same notation from the last property, we now search for similar trajectories where the user minimizes their changes in teleoperation input:
\begin{equation} \label{eq:A2}
    \quad U^* = \text{arg}\min ~ \| \xi - g(U)\|^2 + \lambda \cdot \sum\|u^t - u^{t-1}\|^2
\end{equation}
Solving Equation~(\ref{eq:A2}) outputs trajectories $\xi' = g(U^*)$ where the human inputs are held roughly constant.

\smallskip

\noindent \textbf{Algorithm.} The listed properties produce \textit{counterfactuals}: i.e., what \textit{would have happened} if the human was more noisy, more sparse, or more consistent. In Algorithm~\ref{alg} we add these trajectories to $\choicer$ before leveraging our learning rule from Equation~(\ref{eq:P4}), where $\choicer$ determines the partition function. Overall, Algorithm~\ref{alg} learns from human demonstrations by comparing $\mathcal{D}$ only to a personalized choice set that we are confident the human is capable of inputting.

%% file: simulations.tex
\section{Simulations}

\begin{figure}[t!]
	\begin{center}
		\includegraphics[width=0.95\columnwidth]{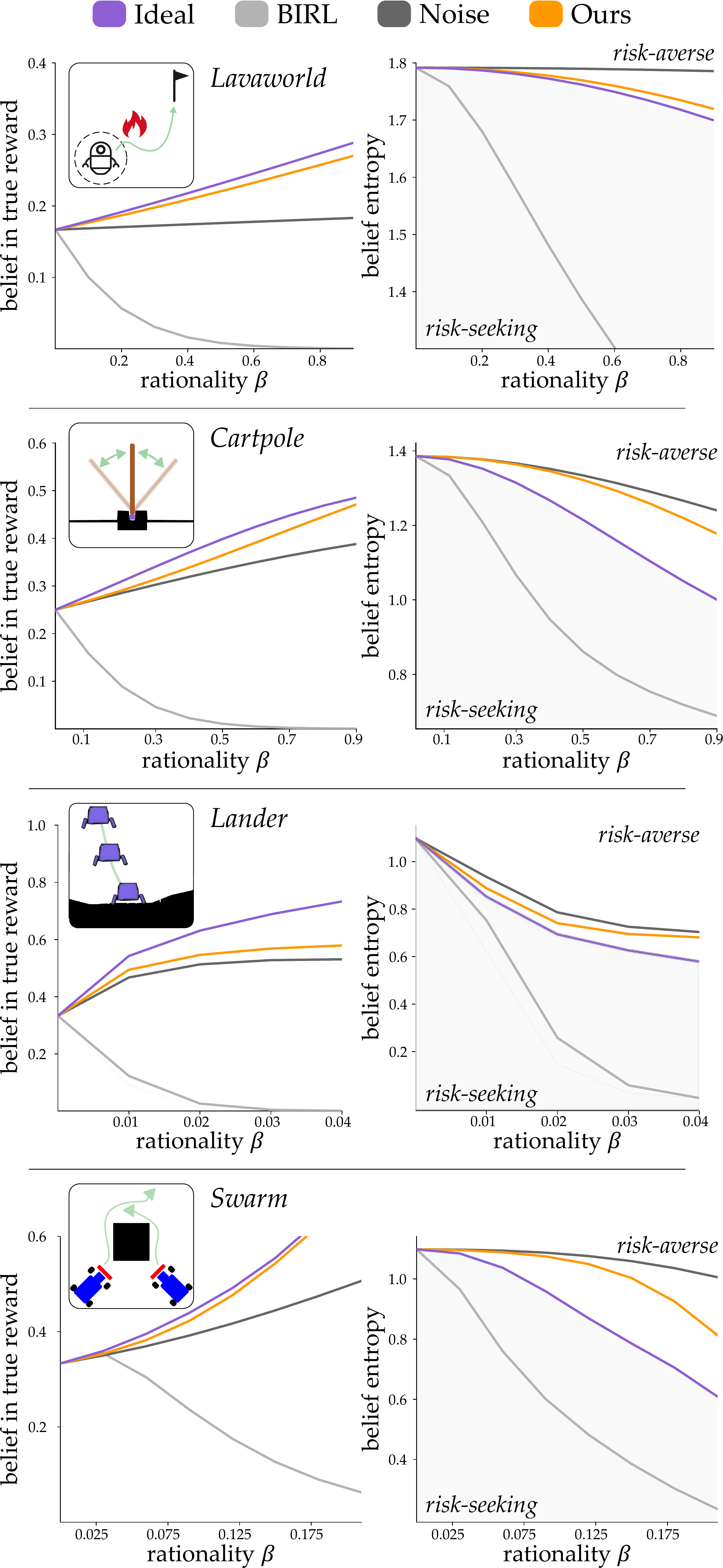}
		\vspace{-0.5em}
		\caption{Simulated users with different levels of rationality ($\beta$) interact in four environments. (Left) The robot's learned belief in the human's true reward. Higher values indicate the robot has learned what the human really meant. (Right) The robot's confidence over its learned belief. Robots with less entropy than \textbf{Ideal} are overconfident in their prediction (risk-seeking), while robots that have more entropy are overly cautious (risk-averse).}
		\label{fig:simulations}
	\end{center}
 	\vspace{-2.em}
\end{figure}

In this section we apply our learning algorithm across a spectrum of simulated humans with different levels of rationality ($\beta$). To better relate these findings to our theoretical results, for now we assume a \textit{discrete} set of possible reward functions (we will move to a continuous hypothesis space in the subsequent user study).

\noindent\textbf{Baselines.}
We compare Algorithm~\ref{alg} (\textbf{Ours}) to three baselines. First we consider a hypothetical world (\textbf{Ideal}) where the robot knows exactly what the human's choice set is ($\choicer = \choiceh$). Next, we test Bayesian inverse reinforcement learning (\textbf{BIRL}) where the robot assumes the human is capable of providing any demonstration \cite{ramachandran2007bayesian}. This leads to $\choicer \supseteq \choiceh$. Finally, we implement a robot that compares human demonstrations to noisy alternatives (\textbf{Noise}). This method is a modified version of D-REX \cite{brown2020better} without ranking feedback, and can be seen as an instance of our formalism where the only counterfactuals are noisy deformations.

\begin{figure*}[t]
	\begin{center}
		\includegraphics[width=1.6\columnwidth]{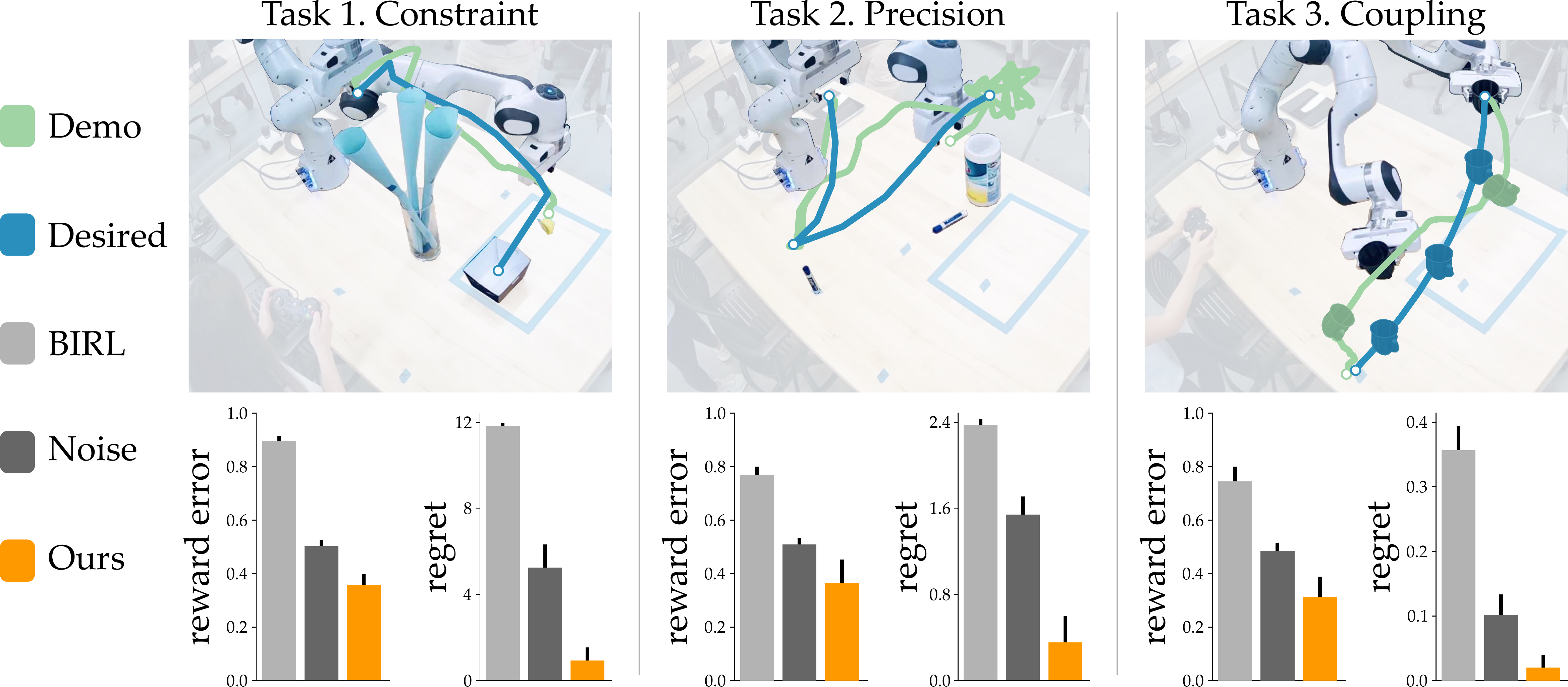}
		\vspace{-0.5em}
		\caption{Human demonstrations and objective results from our user study. (Top) Example user demonstrations and desired behavior for three different tasks. In each task the user is faced with a different challenge that prevents them from inputting their desired behavior. (Bottom) When the robot errs towards towards underestimating the human's capabilities (\textbf{Ours}), it extrapolates what the human really wants from their limited demonstrations. This results in a more accurate estimate of the human's true reward and lower regret across the robot's learned behavior. Error bars indicate standard error about the mean.}
		\label{fig:tasks}
	\end{center}
 	\vspace{-2.em}
\end{figure*}

\noindent\textbf{Environments.} 
We simulate humans with various limitations and test our algorithm across four learning environments (see Fig. \ref{fig:simulations}). In \textit{Lavaworld} the human teleoperates a 2D point mass towards a goal. The human wants to avoid lava, but has a limited range of visibility around the point mass. In \textit{Cartpole} the human tries to keep an inverted pendulum upright, and in \textit{Lander} the human lands a craft on the surface of the moon (OpenAI Gym \cite{BrockmanCPSSTZ16}). Within both \textit{Cartpole} and \textit{Lander} we limit how frequently the simulated human can change their input to mimic the response time of actual users. Finally, in \textit{Swarm} the human navigates an obstacle course with three race cars (developed in Pybullet \cite{coumans2019}). All three cars move simultaneously, but we limit the user to only control one car at a time. For each environment we hand-coded a discrete set of possible reward functions.

\noindent\textbf{Results. }
In Fig. \ref{fig:simulations} we compute the robot's belief using Equation (\ref{eq:P4}) and then evaluate the Shannon entropy of that belief. We find that robots using \textbf{Ours} extrapolate what the human really wanted from their suboptimal demonstrations. Across a spectrum of different environments and different types of simulated users, we also find that robots which overestimate the human's choice set (\textbf{BIRL}) are risk-seeking, while robots which underestimate the human's choice set (\textbf{Noise}, \textbf{Ours}) are risk-averse.

%% file: user-study.tex
\section{User Study}

Motivated by the application of assistive robotics we designed a user study with three manipulation tasks. Participants teleoperated a $7$-DoF robot (Franka Emika) using a $2$-DoF joystick. Users were unable to fully demonstrate what they wanted because of challenges and limitations when controlling the robot (Fig.~\ref{fig:tasks}). Here the robot had a \textit{continuous} space of rewards $r_{\theta}(\xi) = \theta \cdot \Phi(\xi)$, and our proposed approach leveraged Algortihm~\ref{alg} with Metropolis–Hastings sampling to estimate the human's true reward weights $\theta \in \mathbb{R}^k$.

\smallskip

\noindent\textbf{Experimental Setup.}
Each participant attempted to demonstrate three manipulation tasks. In \textit{Constraint}, users needed to drop-off trash in a waste bin. The bin was placed behind an obstacle that the robot must rotate around, but users could only control the end-effector's position. In \textit{Precision}, users needed to carefully stack a marker. This was challenging because noise was injected into the robot's motion. Finally, \textit{Coupling} is our motivational task from Fig. \ref{fig:front}, where users teach the robot to carry coffee upright. Here the robot's translation was coupled with its rotation, so that moving the robot across the table caused the cup to inadvertently tilt.

\smallskip

\noindent\textbf{Independent Variables. }
Participants first provided 3-5 teleoperated demonstrations $\mathcal{D}$ for each task. We then compared what the robot learned with three different methods: \textbf{BIRL} \cite{ramachandran2007bayesian}, \textbf{Noise} \cite{brown2020better}, and \textbf{Ours} (Algorithm~\ref{alg}). \textbf{BIRL} assumes the human can provide any demonstration, while \textbf{Noise} and \textbf{Ours} err towards underestimating the human's capabilities.

\smallskip

\noindent\textbf{Dependent Measures -- Objective.}
We obtained the \textit{Error} $\|\theta - \hat{\theta}\|$ between the true reward weights $\theta$ and the mean of the estimated reward weights $\hat{\theta}$. We also computed \textit{Regret}: $r_{\theta}(\xi^*) - r_{\theta}(\xi_{\mathcal{R}})$. Here $\xi^*$ is the trajectory that maximizes $r_{\theta}$, and $\xi_{\mathcal{R}}$ is the robot's learned trajectory which maximizes $r_{\hat{\theta}}$. If the robot learns what the human really wants, $\xi_{\mathcal{R}} = \xi^*$.

\smallskip

\begin{figure}[t]
    \vspace{0.5em}
	\begin{center}
		\includegraphics[width=0.9\columnwidth]{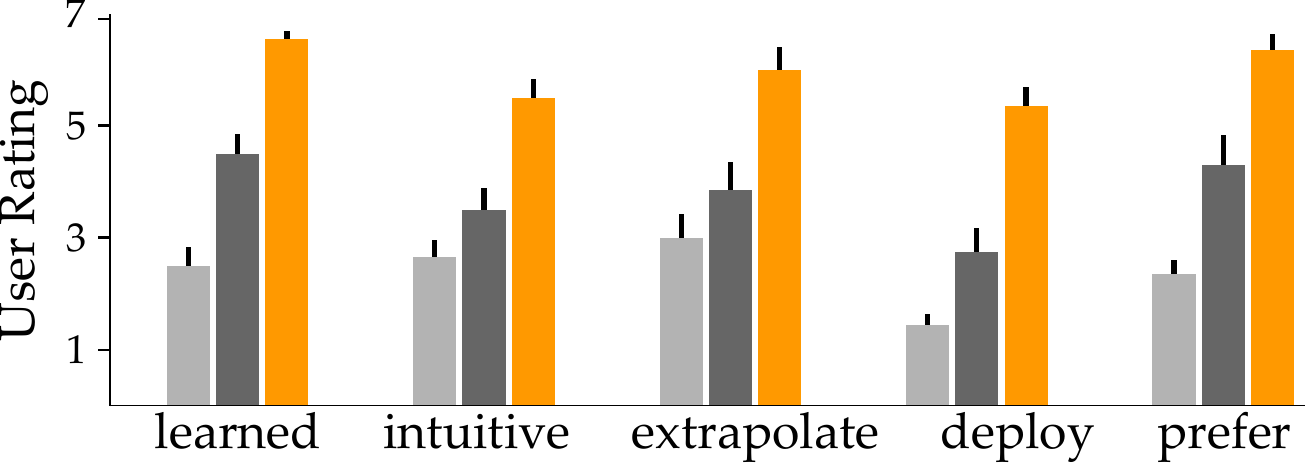}
		\vspace{-0.5em}
		\caption{Survey results from our user study. Higher ratings indicate user agreement. Posthoc comparisons revealed that the differences between our method and the baselines were statistically significant (p < .01).}
		\label{fig:likert}
	\end{center}
 	\vspace{-2.em}
\end{figure}

\noindent\textbf{Dependent Measures -- Subjective.} 
We administered a 7-point Likert scale survey after showing users what the robot learned (see Fig.~\ref{fig:likert}). Questions were organized along five scales: how confident users were that the robot \textit{Learned} their objective, how \textit{Intuitive} the robot's behavior was, whether the robot \textit{Extrapolated} from their demonstrations, how trustworthy users thought the robot was (\textit{Deploy}), and whether they would use the shown method again (\textit{Prefer}).

\smallskip

\noindent\textbf{Participants and Procedure.}
We recruited $10$ subjects from the Virginia Tech student body to participate in our study ($4$ female, average age $22\pm3$ years). All subjects provided informed written consent prior to the experiment. We used a within-subjects design, where we counterbalanced the order of learning algorithms.
\smallskip

\noindent\textbf{Hypothesis. }
\textit{When human teachers face limitations, robots that learn from similar and simpler alternatives best extract what the human wants, and are preferred by users.}
\smallskip

\noindent\textbf{Results -- Objective.}
Our objective results for each task are displayed in Fig. \ref{fig:tasks}. Lower errors indicate the robot's reward estimate approaches the true reward, while lower regret shows that the robot's resulting behavior matches the desired trajectory. With \textbf{BIRL}, the robot learned to replicate what the human demonstrated. But because of their limitations, users were unable to demonstrate their desired trajectory --- and thus \textbf{BIRL} learned suboptimal behavior.

Under \textbf{Noise}, the robot compared the human's demonstrations to similar alternatives (i.e. noisy perturbations). But just considering similar alternatives was not enough to reveal what the human meant. \textbf{Ours} outperformed \textbf{Noise} because it reasoned over both similar and simpler trajectories. Equipped with this choice set, \textbf{Ours} inferred the human's reward (low error), and output near optimal trajectories (low regret).

\smallskip

\noindent\textbf{Results -- Subjective.}
We report the results of the user survey in Fig. \ref{fig:likert}. We received comments such as ``\textit{Method C (\textbf{Ours}) was by far the best and the only one I would trust},'' which highlight the perceived benefits of our approach.

%% file: conclusions.tex
\section{Conclusion}

We analyzed the consequences of over- and underestimating the human's capabilities when learning rewards from demonstrations. Both our risk-sensitivity analysis and experimental results suggest that erring towards underestimating the human's choice set results in safer and more inclusive learning.
\textbf{Limitations:} we recognize that our properties for generating similar and simpler alternatives require hyperparameter tuning, which may depend on the environment.